\documentclass{article} 
\usepackage{arxiv,times}

\usepackage{amsmath,amsfonts,bm}

\def\eqref#1{equation~\ref{#1}}

\def\1{\bm{1}}

\DeclareMathAlphabet{\mathsfit}{\encodingdefault}{\sfdefault}{m}{sl}
\SetMathAlphabet{\mathsfit}{bold}{\encodingdefault}{\sfdefault}{bx}{n}

\usepackage[utf8]{inputenc} 
\usepackage[T1]{fontenc}    
\usepackage{hyperref}       
\usepackage{url}            
\usepackage{booktabs}       
\usepackage{amsfonts}       
\usepackage{nicefrac}       
\usepackage{microtype}      
\usepackage{xcolor}         

\usepackage{amsmath}
\usepackage{graphicx}
\usepackage{multirow}
\usepackage{subcaption}
\usepackage{amssymb}
\usepackage{enumerate,enumitem}
\usepackage{wrapfig}
\usepackage{algorithm, algpseudocode}

\usepackage{setspace}
\usepackage{caption}
\usepackage{colortbl}

\def\mb{\mathbf}
\def\mbb{\mathbb}
\def\mc{\mathcal}

\title{Relational Semantic Grounding for Test-Time Discovery}

\author{
  Tianle Liu\textsuperscript{1}, Gongbo Cheng\textsuperscript{1}, Chenggong Ni\textsuperscript{2}, Minming Gu\textsuperscript{1}, \textbf{Tao Zhou}\textsuperscript{\textbf{3}},\\ ~\textbf{Fuyuan Hu}\textsuperscript{$*$ \textbf{1}}, \textbf{Fan Lyu}\thanks{Corresponding authors.}~~\textsuperscript{\textbf{4}}\\
  \textsuperscript{1}School of Electronics and Information Engineering, Suzhou University of Science and Technology\\
  \textsuperscript{2}School of Intelligence Science and Technology, Nanjing University \\
  \textsuperscript{3}School of Computer Science Engineering, North Minzu University\\
  \textsuperscript{4}Computer Vision Center, Universitat Autònoma de Barcelona\\
  \texttt{tianleliu@post.usts.edu.cn}\\
  \texttt{fanlyu@cvc.uab.cat} ~~~~~~
  \texttt{fuyuanhu@mail.usts.edu.cn}
}

\iclrfinalcopy 
\begin{document}

\maketitle

\begin{abstract}
Test-Time Discovery (TTD) aims to recognize known classes and detect novel ones from a continuous unlabeled test stream after deployment.
Existing TTD methods mainly rely on visual clustering or prototype comparison, and can only assign anonymous labels to discovered classes, leaving them difficult to interpret in open-world applications.
In this paper, we propose SEGA, a Semantic Grounding with Attributes framework for TTD.
SEGA leverages large language models to generate attribute-rich descriptions for known classes and constructs an attribute-anchored visual-semantic space during training.
At test time, we introduce Relational Semantic Grounding (RSG), which grounds unknown samples with similar and dissimilar known-class anchors, prompts the LLM to generate candidate concepts, and reranks them in the learned semantic space.
The selected concept is further used for confidence-guided test-time semantic stabilization.
By coupling attribute-level semantic alignment with relational concept grounding, SEGA transforms anonymous test-time clusters into interpretable semantic concepts.
Experiments show that SEGA improves TTD baselines with the promise of semantic grounding for open-world.
Our code will is available at \url{https://github.com/LeTianL-TT/SEGA-for-TTD}.

\end{abstract}

\section{Introduction}
\label{sec:intro}
\begin{figure}[t]
    \centering    
    \includegraphics[width=1\linewidth]{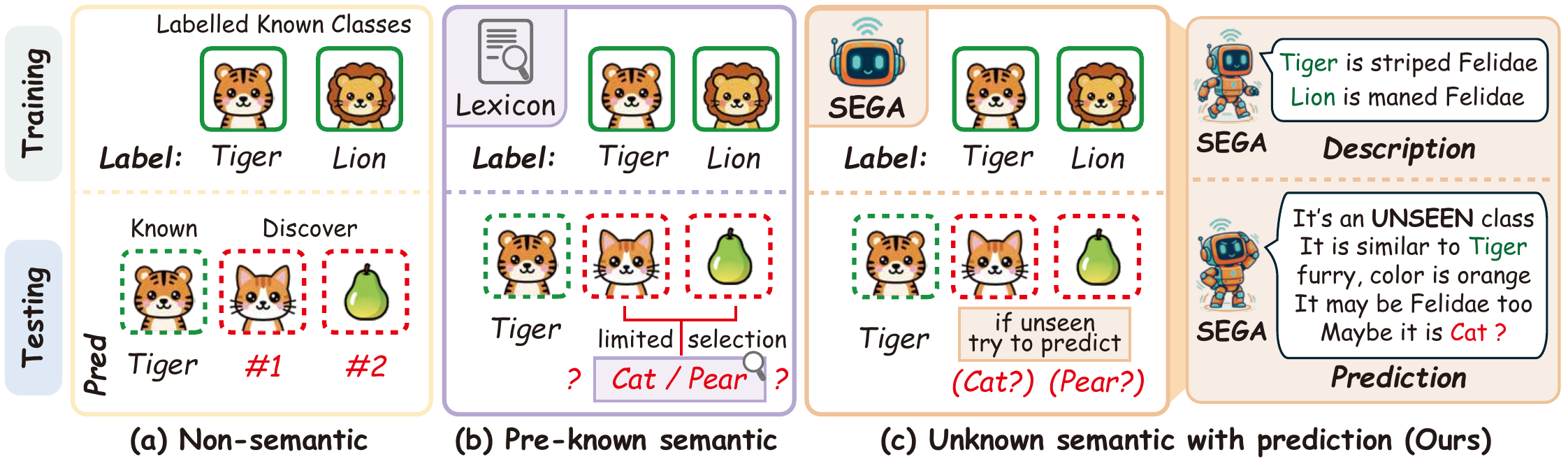}
    \vspace{-10px}
    \caption{
    Comparison among different discovery method. (a) Non-semantic discovery, where unknown samples are detected or clustered online but are only assigned anonymous placeholder labels.
(b) Pre-known semantic discovery handles unknown sample by matching visual features to a predefined category vocabulary, but it is unrealistic and its discovery ability is limited.
(c) Our method aims to more meaningful and open-ended test-time discovery where unknown classes are grounded with visual and semantic relations to known classes, and interpreted through concept generation.
    }
    \label{fig:setting}
    \vspace{-10px}
\end{figure}

Novel Class Discovery (NCD)~\citep{pu2023dynamic, wu2023metagcd} aims to discover unseen categories by transferring knowledge from labeled known classes to an unlabeled data pool. 
However, existing NCD methods assume that the unlabeled data are available in advance, making discovery an offline process. 
This assumption becomes restrictive in open-world deployment, where new categories may continuously emerge in applications. 
Test-Time Discovery (TTD)~\citep{lyu2025test} addresses this setting by requiring the model to recognize known classes, discover novel ones, and update its category space from a continual unlabeled test stream during inference.
In traditional NCD, unknown classes are discovered offline and represented by semantic-agnostic cluster IDs (such as ``\#1''), making immediate semantic interpretation less necessary.
TTD, however, discovers novel classes during deployment, where the discovered classes should be understandable and reusable for humans or subsequent inference rather than merely separated as new clusters.
Existing TTD methods~\citep{lyu2025test, liudaa} largely inherit the NCD-style treatment, assigning semantic-agnostic IDs to unknown classes rather than grounding them as meaningful concepts, which limits their practical value (Fig.~\ref{fig:setting}a).
Some semantic-guided NCD methods~\citep{wang2024semantic, zheng2024textual} introduce textual information through open-vocabulary recognition or fixed lexicon retrieval (Fig.~\ref{fig:setting}b). 
However, these strategies assume that candidate novel categories can be covered by a predefined vocabulary, which is misaligned with open-world TTD where possible novel classes are unknown before deployment. 
A simple alternative is to use vision-language models with class-name prompts such as ``a photo of a [class]'', but such prompts provide only coarse semantic anchors and lack fine-grained attributes for relating unknown classes to known ones~\citep{tan2026logic, mitigateocl,fewshotcil}. 
LLMs, with their rich conceptual and attribute-level knowledge, offer a more flexible source of semantic cues for open-ended discovery~\citep{du2024ipo,fu2025llmdet}. 
This suggests that attribute-rich descriptions generated by LLMs can provide a semantic basis for TTD, allowing known-class attributes to serve as transferable bridges for grounding novel classes at test time.

In this paper, we propose \textbf{Semantic Grounding with Attributes (SEGA)}, a framework that transforms TTD from anonymous clustering into semantic concept generation (Fig.~\ref{fig:setting}c). 
During training, SEGA first uses LLM-generated attribute descriptions to enrich known classes and learns an attribute-aware visual-semantic space through cross-modal alignment. 
At test time, when an unknown-class candidate is detected, we introduce Relational Semantic Grounding (RSG) to describe it by its relations to known-class anchors and visual attributes, and then query an LLM to generate candidate category names. 
Since free-form LLM generation can produce overly broad or implausible names, SEGA reranks the candidates according to their plausibility in the learned semantic space. 
The selected concept is further converted into a pseudo description for confidence-guided test-time training, which stabilizes subsequent discovery and reduces the effect of noisy semantic predictions. 
By coupling attribute-level semantic alignment, relational concept generation, and semantic stabilization, SEGA provides an interpretable TTD pipeline that grounds discovered unknown classes as meaningful concepts rather than anonymous tokens. 
Experiments on four TTD benchmarks show that SEGA improves discovery performance over recent TTD baselines, with up to 4.55\% gain in agreement metrics.
In summary, our contributions are: 

\begin{enumerate}[label=(\arabic*),left=0pt,itemsep=0pt,parsep=0pt]
\vspace{-5px}
\item We identify and address the semantic grounding problem in TTD, where discovered unknown classes should be represented as meaningful concepts rather than anonymous cluster IDs, and propose SEGA as a unified framework for semantic test-time discovery.
\item We introduce RSG, which combines LLM-generated known-class attributes, relational semantic anchors, and semantic reranking to generate reliable candidate names for unknown classes.
\item We evaluate SEGA on four TTD benchmarks, showing improvements over prior TTD methods in discovery performance, known-class stability, and semantic quality of the generated class names.
\end{enumerate}

\section{Related Work}
\label{sec:related}
\textbf{From Semantics-Agnostic to Semantic-Guided Category Discovery.}~\citep{han2020automatically, zhang2023promptcal, zhong2021neighborhood} Novel Class Discovery (NCD) aims to identify new categories from unlabeled data by transferring knowledge from known classes.
GCD~\citep{vaze2022generalized} and PromptCCD~\citep{cendra2025promptccd} replace the traditional k-means stage with contrastive learning prompting for purer prototypes. On-the-fly Category Discovery (OCD)~\citep{du2023onfly, zheng2024PHE} pushes NCD toward real-time operation by hashing images and assigning them according to Hamming-distance.
Later, some work start to explore incorporating textual information into category discovery to address the semantics gap. 
SNCD~\citep{wang2024semantic} cluster unlabeled images but also semantically match them by constructing semantic similarity matrices. Some CLIP-guided GCD works~\citep{zheng2024textual,wang2025get} leverage pre-trained vision-language alignment to guide discovery with semantic priors. And the open-vocabulary paradigm~\citep{wu2024openvoca,zhu2024openvoca} has further expanded the semantic horizon. However, these approaches operate in offline or static settings and depend on predefined or retrieved textual descriptions. 

\noindent
\textbf{Test-Time Discovery (TTD)} requires models to detect and learn novel classes in real-time as samples arrive sequentially. The pioneering HM~\citep{lyu2025test} proposes a training-free approach using locality-sensitive hashing for efficient sample retrieval and prototype comparison. DAA~\citep{liudaa} introduces a lightweight trainable adapter to amplify feature-level discrepancies between known and unknown classes. While these methods achieve impressive performance in visual feature discrimination, discovered classes are assigned placeholder labels without any interpretable meaning.

\noindent
\textbf{LLM for Visual Knowledge.} LLM-generated descriptions for visual recognition are widely used in different work recently~\citep{zhang2025attribute,lei2025open,fu2025llmdet}. Detailed visual descriptions are revealed to outperform fixed templates and can improve task like zero-shot, open-vocabulary and attribute-concept annotation~\citep{ding2026vd, myyicme, tan2026towards}.
The dynamic, streaming nature of TTD with its real-time constraints and irreversible decisions remains unexplored in the context of LLM-assisted semantic discovery~\citep{lyu2025mitigating,ni2025maintaining,Tan2026iclr}. Our work bridges this gap by enabling test-time LLM reasoning for semantic interpretation, transforming TTD from visual clustering to conceptual understanding.

\section{Method}
\label{sec:method}
\subsection{TTD and a CLIP-based baseline}
TTD aims to recognize known classes and discover unknown ones in a continuous unlabeled stream after model deployment.
Formally, let \(\mathcal{D}_{\text{train}} = \{(x_i, y_i)\}_{i=1}^{N}\) denote the training set, where \(x_i \in \mathcal{X}\) is input image, and \(y_i \in \mathcal{Y}_{\text{kn}}\) is its number label from the known class set \(\mathcal{Y}_{{kn}}\). 
During testing, the model encounters a test stream \(\mathcal{D}_{\text{test}} = \mathcal{D}_{\text{kn}} \cup \mathcal{D}_{\text{un}}\), where \(\mathcal{D}_{\text{un}}\) may contain unknown classes that were never seen during training. TTD requires the model to satisfy two objectives simultaneously. 
First, \textit{novel class discovery and learning}, where inputs \( x \in \mathcal{D}_{\text{test}}^{\text{un}} \) must be identified as belonging to unseen classes and assigned new labels to distinguish them from known-class samples.
Second, \textit{unified classification}, where all inputs \( x \in \mathcal{D}_{\text{test}} \) are accurately classified into either a known class \( y \in \mathcal{Y}_{\text{kn}} \) or a novel class \( y \in \mathcal{Y}_{\text{seen}} \), with \( \mathcal{Y}_{\text{seen}} \) denoting the set of dynamically discovered novel classes.

Inspired by recent vision-language discovery methods~\citep{zheng2024textual,wang2025get}, we adopt CLIP~\citep{radford2021clip} as the backbone.
To preserve linguistic knowledge, we freeze the pretrained text encoder and fine-tune the image encoder, with lightweight learnable projection layers attached to both modalities. 
The projection layers map image and text features into a shared attribute-aware space and are updated during training and test-time adaptation, enabling flexible alignment without modifying the frozen text encoder.

\begin{wrapfigure}[14]{r}{0.6\textwidth}
    \centering
    \vspace{-10px}
    \includegraphics[width=\linewidth]{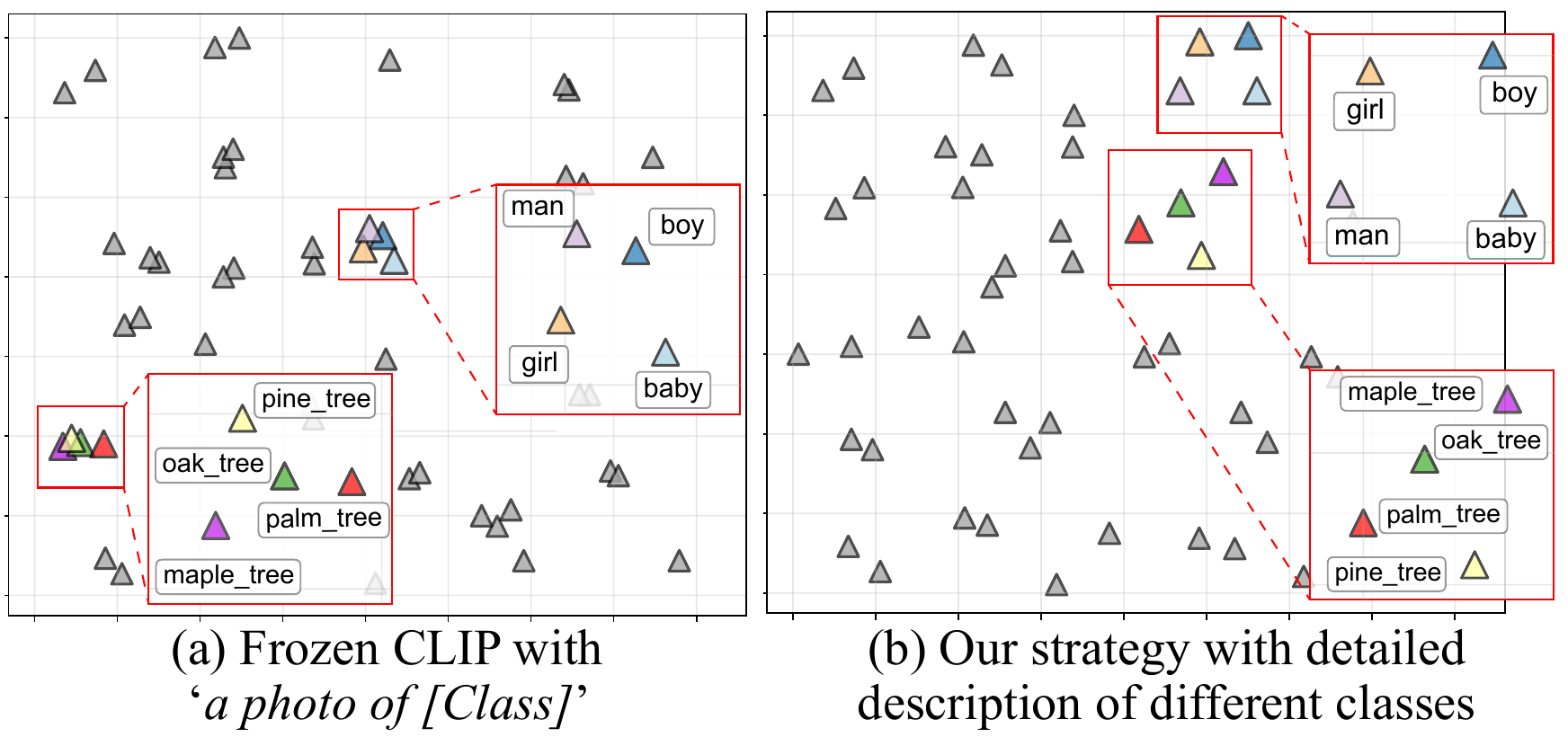}
    \vspace{-10px}
    \caption{T-SNE visualization of the CLIP-based textual feature for different kinds of text prompt. 
    }
    \label{fig:clipvsllm}
\end{wrapfigure}

Standard CLIP-based methods often use class-name prompts such as \textit{``a photo of [CLASS]''}, which provide only coarse semantic anchors. 
As shown in Fig.~\ref{fig:clipvsllm}, such template-based text embeddings are dominated by class names and may entangle semantically related but visually distinct categories. 
In the following subsection, we instead enrich each known class with LLM-generated attribute descriptions, following the intuition that attributes provide transferable semantic cues across categories~\citep{du2024ipo,zheng2024textual}. 
Attribute descriptions capture fine-grained cues such as shape, color, texture, and parts, helping distinguish similar classes while preserving their semantic relations.

\subsection{Attribute-Anchored Semantic Space Construction}

Based on the above analysis, we first construct an attribute-anchored visual-semantic space using known-class training data to support semantic grounding at test time.
For each known class \(c\in\mathcal{Y}_{\mathrm{kn}}\), we query an LLM, e.g., GPT-4o-mini, to generate an attribute-rich description \(t_c\) that summarizes discriminative visual properties such as color, shape, texture, and parts\footnote{
\;Such as "Describe [CLASS] in detail, including visual attributes like color, shape...".
}.
This yields a set of textual anchors \(\mathcal{T}=\{t_c\}_{c\in\mathcal{Y}_{\mathrm{kn}}}\), which provide semantic references for known classes\footnote{
\;"A zebra has black and white stripes, a horse-like shape..."
}.
The text encoder is frozen to preserve its linguistic knowledge, while the image encoder and lightweight projection layers are trainable.
Given an image \(x\) and the textual anchor \(t_{c}\) of its class, we obtain
\begin{equation}
    \mathbf{z_v}(x) = P_v\bigl(E_v(x); W_v\bigr)
    \quad
    \mathbf{z_t}(t_c) = P_t\bigl(E_t(t_c); W_t\bigr),
\end{equation}
where \(P_v\) and \(P_t\) map both modalities into a shared attribute space \(\mathcal{Z}\).

\textbf{Semantic anchor calibration.} 
The LLM-generated descriptions are not used as ordinary class labels. Actually, they act as semantic anchors \(\mathbf{a}_c=\mathbf{z}_t(t_c)\) , and all known-class anchors form an anchored textual manifold 
\(\mathcal{M}_t=\{\mathbf{a}_c\mid c\in\mathcal{Y}_{\mathrm{kn}}\}\) that define how known classes are arranged in the attribute space.
However, directly using descriptions as anchors can be unstable, since pretrained models are not inherently arranged for rich descriptions.
We therefore \textit{calibrate} these textual anchors so that descriptions of the same-class textual neighbors 
\(\mathcal{N}_t^+(i)=\{j\in\mathcal{B}:y_j=y_i\}\) collapse into compact semantic regions, while anchors from different classes 
\(\mathcal{N}_t^-(i)=\{j\in\mathcal{B}:y_j\neq y_i\}\) occupy separable territories.
A well-calibrated anchor space therefore provides a reliable coordinate system for subsequent relational inference, rather than merely serving as a set of text classifiers.
\\
Specifically, given a test batch $\mc{B}$, we ensure that the text embedding space preserves class-level discriminability with attribute-rich descriptions by

\begin{equation}
\mathcal{L}_{\text{text}} = -\frac{1}{|\mc{B}|} \sum_{i=1}^{|\mc{B}|}\log \frac{
\sum_{j: y_j = y_i} \exp[\mathbf{z_t}({t_i})^\top\mathbf{z_t}({t_j})]}{
\sum_{{c}\in \mc{B}} \exp[\mathbf{z_t}({t_i})^\top \mathbf{z_t}({t_c})]
}.
\end{equation}

\begin{figure*}[t]
    \centering        \includegraphics[width=\linewidth]{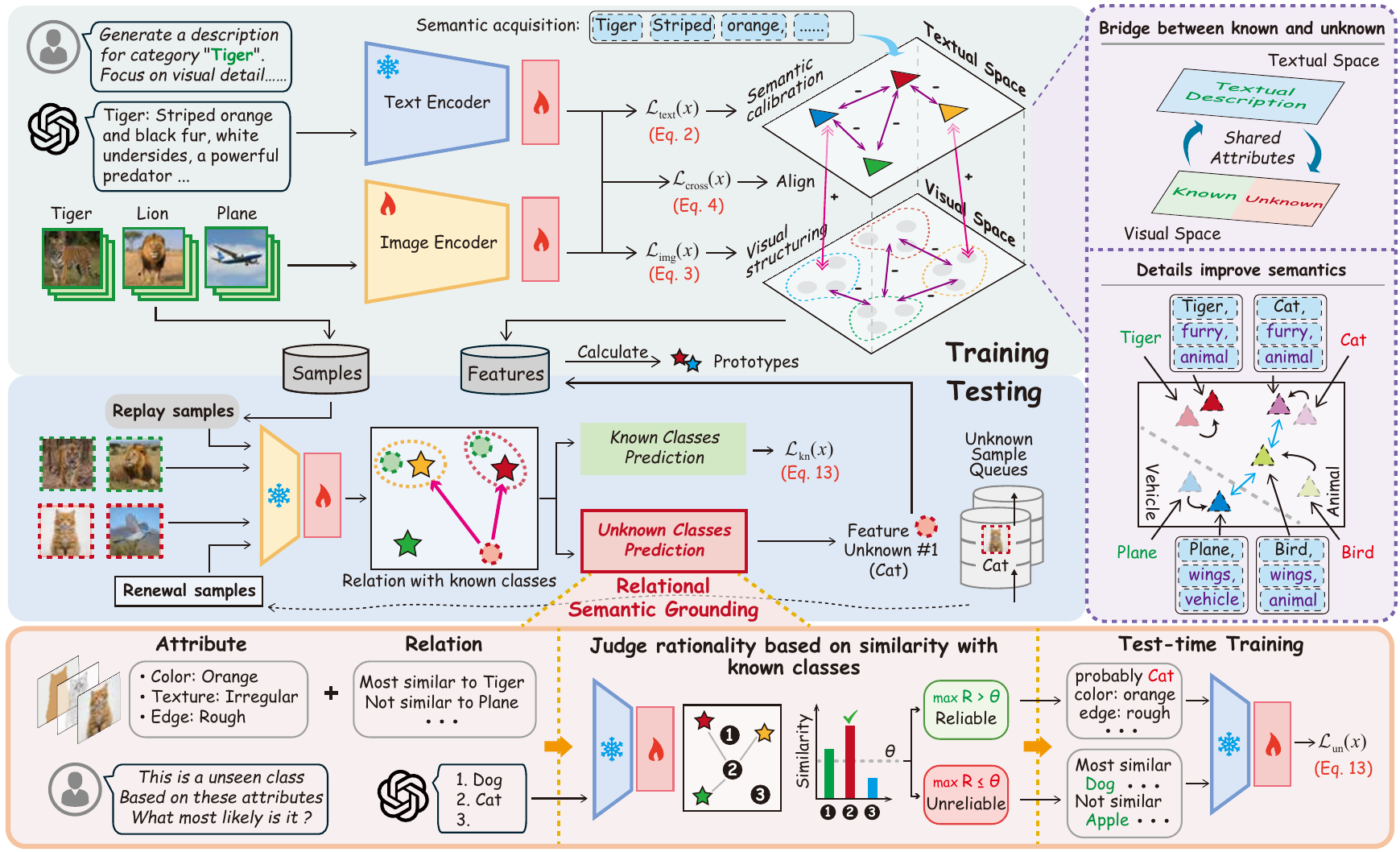}
    \vspace{-18px}
    \caption{
     Method Schema. During training, LLM-generated attribute descriptions are used to construct a semantic space and guide cross-modal alignment between images and text. During testing, we discovers unknown classes, retrieves their semantic relations to known class anchors, and queries the LLM to generate candidate category names. The candidates are then grounded and filtered by the RSG module, and the selected prediction is used to guide confidence-aware test-time updates.
    }
    \label{fig:method}
    \vspace{-15px}
\end{figure*}

\vspace{-3px}
\textbf{Visual neighborhood structuring.} While \(\mathcal{M}_t\) provide the semantic coordinate system, the visual stream also needs a stable neighborhood structure before it can support discovery.
Let \(\mathbf{v}_i=\mathbf{z}_i(x_i)\)  denote the visual feature of image \(x_i\) and if visual manifolds \(\mathcal{M}_v=\{\mathbf{v}_i\}\) are scattered or overly fitted to known-class decision boundaries, an unseen sample may be absorbed by a known class or isolated as noise.
We therefore organize \(\mathcal{M}_v\)  with a dual objective by
\begin{equation}
\mathcal{L}_{\text{img}} = -\frac{1}{|\mc{B}|} \sum_{i=1}^{|\mc{B}|} \left(\log \frac{
\exp \left[ \mathbf{z_i}(x_i)^\top \mathbf{\tilde{z}_i}(x_i) \right]
}{
\sum_{{c}\in \mc{B}} \exp\left[\mathbf{z_i}(x_i)^\top \mathbf{{z}_i}(x_c)
\right]}
+
\log \frac{ \sum_{j: y_j = y_i}
\exp \left[ \mathbf{z_i}(x_i)^\top \mathbf{z_i}(x_j)\right]
}{
\sum_{c \in \mc{B}} \exp\left[\mathbf{z_i}(x_i)^\top \mathbf{z_i}(x_c)\right]}\right),
\end{equation}
where \(\mathbf{\tilde{z}_i}\) denotes the feature of an augmented view of \(x_i\). In this way, the visual representation is made robust to visual variations and the image encoder learns a visual manifold that is suitable for both recognizing known classes and detecting deviations caused by emerging unknown classes.

\textbf{Cross-modal coordinate binding.} The calibrated textual manifold \(\mathcal{M}_t\) and the structured visual manifold \(\mathcal{M}_v\) are useful only if they share the same coordinate system. 
To bridge this gap, we build image-anchor correspondences 
\(\mathcal{G}=\{(\mathbf{v}_i,\mathbf{a}_{y_i})\}_{i\in\mathcal{B}}\), where each visual feature is paired with the semantic anchor of its class.
Therefore, we further introduce a cross-modal binding objective along \(\mathcal{G}\) connecting each image to its corresponding attribute-rich textual anchor without which the image encoder may rely on spurious visual correlations.
This binding encourages visual features to inherit the semantic directions encoded by the LLM descriptions, such as shape, texture, color, and part-level cues.
Once bound, visual representations are no longer determined purely by appearance and unknown sample can then be described through its position relative to known semantic anchors because they share the same semantic coordinates in the attribute space. 
This binding is achieved via:
\begin{equation}
\mathcal{L}_{\text{cross}} = -\frac{1}{2|\mc{B}|} \sum_{i=1}^{|\mc{B}|} \left(
\log \frac{
\exp \left[ \mathbf{z_i}(x_i)^\top \mathbf{z_t}(t_i)\right]
}{
\sum_{{c}\in \mc{B}}  \exp \left[\mathbf{z_i}(x_i)^\top \mathbf{z_t}(t_c)\right]
} + 
\log \frac{
\exp \left[\mathbf{z_t}(t_i)^\top \mathbf{z_i}(x_i)\right]
}{
\sum_{{c}\in \mc{B}}  \exp \left[\mathbf{z_t}(t_i)^\top \mathbf{z_i}(x_c)\right]
}
\right).
\end{equation}

\vspace{-5px}
The three objectives play complementary roles which together prepare the model for testing where known classes become stable semantic references, while unknown samples can be interpreted by their relational position in the anchored visual-semantic space.
The total training loss is:
\begin{equation}
\mathcal{L}_{\text{total}} =  \mathcal{L}_{\text{text}} + \mathcal{L}_{\text{img}} +  \mathcal{L}_{\text{cross}}.
\end{equation}

\subsection{Relational Concept-Level Semantic Inference}
\label{subsec:joint}
When encountering the test sample $x$ with a known or unknown class, the prediction procedure involves calculating the cosine similarity $s_k=\text{sim}(\mb{z_i}(x), \mathbf{p}_c)$ between the image feature $\mb{z_i}(x)$ and the prototypes $\mathbf{p}$ of known classes to determine their classification. 
Based on the max similarity scores, we apply a confidence threshold \(\gamma\) to determine whether a test sample $x$ belongs to a seen class or an unseen class.
\begin{equation}
\hat{y} = \begin{cases} 
\text{arg max}_{c\in(\mc{Y}_{\text{kn}}\cup\mc{Y}_{\text{seen}})}(s_k), & \text{if } \max_c s_k > \gamma, \\
\text{new discovered class}, & \text{otherwise}.
\end{cases}
\end{equation}

\textbf{Prompt construction and candidate generation.} If the sample is classified as an unseen class \(\hat{y} \in {\mathcal{Y}}_\text{unseen}\), two sets Top-$k$ most similar classes $\mathcal{S}_{\text{sim}}$ and Top-$k$ least similar classes $\mathcal{S}_{\text{dis}}$ are extracted. These similarity patterns serve as relational anchors, characterizing the unknown class by what it resembles and what it differs from in the known semantic space.
\begin{equation}
\mc{S}_{\text{sim}} = \{(c, s_k) \mid c \in \text{top-}k \text{ of } s_k\} \qquad
\mc{S}_{\text{dis}} = \{(c, s_k) \mid c \in \text{bottom-}k \text{ of } s_k\}
\end{equation}

Relational similarity alone may be insufficient for fine-grained naming. To provide additional discriminative cues, we extract interpretable visual attributes $\mathcal{A}$ from the image $x$, which include color statistics (dominant hue, saturation, temperature), texture properties (roughness, regularity), lighting conditions (brightness, uniformity), and structural characteristics (edge density, composition). These attributes are computed using only low‑level operations on the raw images.
\begin{equation}
\mathcal{A} = \Phi_{\text{attr}}(\mathbf{x}) = \{\mathcal{A}_{\text{color}}, \mathcal{A}_{\text{texture}}, \mathcal{A}_{\text{lighting}}, \mathcal{A}_{\text{structure}}\}
\end{equation}

Then we construct a prompt $\mathcal{P}=(\mathcal{S}_{\text{sim}}, \mathcal{S}_{\text{dis}}, \mathcal{A})$ that integrates both relational and attribute information to instruct the LLM to generate $M$ candidate category names for the discovered class:
\begin{equation}
\mathcal{C}_{\text{LLM}} = \{\hat{c}_1, \hat{c}_2, \ldots, \hat{c}_M\} = \text{LLM}(\mathcal{P}{(\mathcal{S}_{\text{sim}}, \mathcal{S}_{\text{dis}}, \mathcal{A})})
\end{equation}

\textbf{Semantic reranking with text encoder.} Some of these candidates are plausible but potentially noisy. To ground the LLM's free-form generation in the actual visual-semantic space, we propose a semantic reranking mechanism. We encode both the LLM-generated candidates and the top-$k$ most similar known classes with our text encoder. And the relevance score $R$ for each candidate $\hat{c}$ is computed as the average similarity with the known reference classes.
\begin{equation}
R(\hat{c}) = \frac{1}{|\mathcal{S}_{\text{sim}}|} \sum_{c \in \mathcal{S}_{\text{sim}}} \mathbf{z_t}(\hat{c}) \cdot \mathbf{z_t}({y_c}),\quad\hat{c} \in \mathcal{C}_{\text{LLM}}
\end{equation}

The top-ranked candidate serves as the semantic label $\hat{y}_t$ for the discovered class, and this class will become a seen class \({\mathcal{Y}}_\text{seen} \leftarrow \hat{y} \),
and the model will treat this sample as the first instance of this new seen class. The image feature \(\mb{z_i}(x)\) will be used as the initial prototype for this class, denoted as \(\mathbf{p}_{\hat{y}}\). Subsequent samples will be compared to all known and seen prototypes, including this new \(\mathbf{p}_{\hat{y}}\).
\begin{equation}
\hat{y}_t = \arg\max_{\hat{c} \in \mathcal{C}_{\text{LLM}}} R(\hat{c})
\end{equation}


\subsection{Confidence-Guided Test-Time Semantic Stabilization}  
\label{sec:test_time_train}
We acknowledge that precisely predicting the ground-truth label of a never-before-seen class is inherently challenging due to the open-world nature of the task. By introducing LLMs, we take an initial step toward addressing the lack of semantic interpretability which is a fundamental challenge intrinsic to TTD. However, we believe that while accurate label prediction is the ultimate goal for downstream targets, it is not entirely strict for auxiliary classification. What matters more is that the discovered semantic label $\hat{y}_t$ occupies a reasonable position in the text embedding space.

Therefore, we propose a reliability strategy for test-time training guided by the semantic reranking score \(R(\hat{c})\). A high \(R\) indicates that the top candidate \(\hat{y}_t\) is semantically consistent with known reference classes, suggesting a stable prediction. Conversely, a low \(R\) implies ambiguous generation. For a sample \(x\) with a stable prediction, we construct an attribute-rich textual description \(\tilde{t}_{\hat{y}}\) with the predicted label \(\hat{y}_t\). While a sample with an uncertain prediction, we avoid using a potentially incorrect label. Instead, we construct a relational textual prompt based solely on the similarity set \(\mathcal{S}_{\text{sim}}\) and dissimilarity set \(\mathcal{S}_{\text{dis}}\).
\begin{equation}
\tilde{t}_{\hat{y}} = \begin{cases} 
\mathcal{P}(\hat{y}_t,  \mathcal{A}, \mathcal{S}_{\text{sim}}, \mathcal{S}_{\text{dis}}), & \text{if } \max_{\hat{c} \in \mathcal{C}_{\text{LLM}}} R(\hat{c}) > \theta, \\
\mathcal{P}(\mathcal{S}_{\text{sim}}, \mathcal{S}_{\text{dis}}), & \text{otherwise}.
\end{cases}
\end{equation}

For a test batch \(\mathcal{B}\), the first type of loss applies to samples with the pseudo labels:
\begin{equation}
\begin{array}{l}
\mathcal{L}_{\text{kn}}(x,t)=
\mbb{E}_{\substack{\hat{y}\in\mathcal{Y}_{\text{kn}}}}
\Bigl[\,
\mathcal{L}_\text{text}(t\mid\hat{y})
+
\mathcal{L}_\text{img}(x\mid\hat{y})
+
\mathcal{L}_\text{cross}((x,t)\mid\hat{y})
\Bigr],
\\
\mathcal{L}_{\text{un}}(x,\tilde{t}_{\hat{y}}) =  \mbb{E}_{\substack{\hat{y}\in\mathcal{Y}_{\text{un}}}}
\Bigl[\,
\mathcal{L}_\text{text}(\tilde{t}_{\hat{y}} \mid\hat{y})
+
\mathcal{L}_\text{img}(x\mid\hat{y})
+
\mathcal{L}_\text{cross}((x,\tilde{t}_{\hat{y}})\mid\hat{y})
\Bigr].
\end{array}
\end{equation}

The second kind of loss follows DAA~\citep{liudaa} to conduct replay for known classes and renewal for seen classes by retraining on samples stored in a memory bank. Since both initial predictions and early-stage classifications are inevitably error-prone, we incorporate a self-correction mechanism as DAA does to, which in the TTD setting can effectively refine subsequent classification performance. Specifically, we retrieve the samples regularly, and these samples would be reprocessed, re-evaluated, and update SEGA again. The overall testing phase loss is shown below:
\begin{equation}
\begin{array}{c}
\mathcal{L}_\text{test} = \lambda_1\cdot\mbb{E}_{\hat{y}\in\mc{Y}_\text{kn}}
\left[
\mathcal{L}_{\text{kn}}(x,t|\hat{y}) + \mathcal{L}_\text{kn}^{\text{replay}}(x)
\right] 
+ \lambda_2\cdot\mbb{E}_{\hat{y}\in\mc{Y}_\text{seen}}
\left[
\mathcal{L}_{\text{un}}((x,\tilde{t}_{\hat{y}})|\hat{y})
+\mathcal{L}_\text{un}^{\text{renew}}(x|\hat{y})
\right],
\end{array}
\end{equation}
where \(\lambda_1\) and \(\lambda_{\text{2}}\) are hyperparameters that balance the contributions of each loss term.

\section{Experiment}
\label{sec:exp}

\begin{table*}[t]
\centering
\setlength{\tabcolsep}{7pt}
\caption{Comparisons on four datasets. Real-time evaluation reflects the accumulated performance across all test batches, while post evaluation reassesses all test samples after testing. \textbf{Bold} and \underline{underlining} denote the best and second-best results.}
\vspace{-7px}
\resizebox{\textwidth}{!}{
\begin{tabular}{cl|rrr|rrrr}
\toprule
\multirow{2}{*}{} &  \multirow{2}{*}{\textbf{Method}\;\;\;\;} &  \multicolumn{3}{c|}{\textbf{Real-time Evaluation}} &
  \multicolumn{4}{c}{\textbf{Post Evaluation}} \\
  &           &  \multicolumn{1}{c}{KA$\uparrow$} & \multicolumn{1}{c}{TA$\uparrow$} &  \multicolumn{1}{c|}{CA$\uparrow$} &  \multicolumn{1}{c}{KA$\uparrow$} & \multicolumn{1}{c}{TA$\uparrow$} &  \multicolumn{1}{c}{CA$\uparrow$} &  \multicolumn{1}{c}{KF$\downarrow$} \\
\midrule
\multirow{7}{*}{\rotatebox[origin=c]{90}{\textbf{CIFAR100D}}}                                        
&  L2P~\citep{wang2022learning} &  59.93±2.15 &  8.57±2.49 &   43.10±4.30 &   50.53±7.25 &  9.60±1.50 &  27.39±2.11 &   27.85±7.21 \\
& DP~\citep{wang2022dualprompt}        & 66.09±1.01       & 8.80±1.69 & \underline{48.34}±6.78 &  56.19±2.00 & 8.68±2.06 & 28.93±2.25 & 29.06±1.99    \\
& GMP~\citep{cendra2025promptccd}       & 72.77±1.20     & 7.37±\underline{0.88} & 42.26±4.54 &  67.21±2.53 & 13.04±2.59 &  27.10±1.77 & 17.69±2.53  \\
& PHE~\citep{zheng2024PHE}     & 72.35±1.12 & 16.74±1.05 &  37.34±1.31 & 70.12±1.07 & 14.55±\textbf{1.01} & 24.65±1.35  & \textbf{2.12}±0.92  \\
\cmidrule{2-9}
& HM~\citep{lyu2025test}  & 79.17±\textbf{0.13} &  21.13±\textbf{0.62} &  \textbf{56.37}±1.42 &   \underline{80.73}±1.59 & 31.03±1.24 &  34.81±\underline{1.22}  &  3.41±1.49 \\
& DAA~\citep{liudaa}       & \underline{80.81}±\underline{0.44} & \underline{31.94}±0.98  & 45.38±\textbf{0.88}  & 80.27±\textbf{0.60} & \underline{35.63}±\underline{1.11} &  \underline{37.31}±\textbf{1.20}
& 4.10±\underline{0.60}  \\
\rowcolor{blue!10} & Ours      & \textbf{83.49}±0.64 & \textbf{34.88}±0.92 & 47.94±\underline{1.04} & \textbf{82.78}±\underline{0.89} & \textbf{38.64}±1.20 & \textbf{39.15}±1.24 & \underline{3.04}±\textbf{0.54}  \\
\midrule
\multirow{7}{*}{\rotatebox[origin=c]{90}{\textbf{CUB200D}}}        

& L2P~\citep{wang2022learning}       & 46.22±1.53 &  9.01±\underline{0.87} &   55.37±7.79 &    31.97±3.35 &  4.75±\underline{0.73} &   24.48±1.65 &  42.29±3.14  \\
& DP~\citep{wang2022dualprompt}         & 53.69±1.24 &  43.68±2.20 &   45.68±5.88 &  63.37±3.27 &  45.94±0.91 &  7.92±\underline{1.10}&  5.85±3.11  \\
& GMP~\citep{cendra2025promptccd}       &  62.97±1.33 &  46.44±1.87 & 47.99±4.34 &  58.11±3.00 &  48.02±1.20&  10.31±1.45&  5.46±2.77  \\
& PHE~\citep{zheng2024PHE}   &57.29±1.23 & 32.68±1.66 &  57.22±2.75 & 54.77±1.75 & 27.92±\textbf{0.43} & 21.36±1.36 & \textbf{2.18}±0.71  \\
\cmidrule{2-9}
& HM~\citep{lyu2025test}     & 66.20±\underline{0.55} &  \underline{58.30}±2.37 &   43.33±6.10 &  64.42±\underline{0.65} &  \textbf{65.28}±1.78&  37.25±4.90 &    4.07±\underline{0.47}  \\
& DAA~\citep{liudaa}      & \underline{68.09}±\textbf{0.33} & 54.49±1.03 & \textbf{58.34}±\underline{1.35} & \underline{66.26}±\textbf{0.39} & 64.13±2.23 &  \underline{44.65}±\textbf{0.88} & 3.60±\textbf{0.39}   \\
\rowcolor{blue!10} & Ours      &  \textbf{68.68}±1.21 & \textbf{58.30}±\textbf{0.62} & \underline{57.72}±\textbf{1.01} & \textbf{67.73}±0.94 & \underline{64.25}±1.63 & \textbf{46.00}±1.22 & \underline{2.83}±0.62\\

\midrule
\multirow{7}{*}{\rotatebox[origin=c]{90}{\textbf{T-ImageNetD}}}
& L2P~\citep{wang2022learning}       & 46.25±1.41 &  7.79±2.92 &    53.55±6.47 &  29.50±3.77 &  10.38±2.42 &  23.28±\textbf{0.79} &  47.97±3.77  \\
& DP~\citep{wang2022dualprompt}        &46.51±\textbf{0.58} &  6.41±0.93 &  58.27±6.10 & 28.53±3.33 &  9.63±2.10 &  26.80±1.38 &  47.57±3.32   \\
& GMP~\citep{cendra2025promptccd}       &62.47±1.40 &  6.25±1.72 &  58.02±4.29 &  63.95±2.04 &  15.64±2.63 &   26.31±2.33 &    16.86±2.04   \\
& PHE~\citep{zheng2024PHE}  & 63.28±1.29 & 13.54±1.20 & 70.33±3.80 &  58.39±\textbf{1.14} & 12.80±\textbf{1.05} & 19.42±1.45 &  1.88±\textbf{0.35}  \\
\cmidrule{2-9}
& HM~\citep{lyu2025test}      & 75.31±1.31 &  16.04±\underline{0.76} &  \textbf{73.81}±2.67 &   74.94±2.20 &  16.23±\underline{1.24} &  \underline{37.43}±1.30 &  \textbf{1.15}±2.18  \\
& DAA~\citep{liudaa}      & \underline{76.38}±\underline{0.82} & \underline{24.10}±0.80 &  71.70±\underline{2.02} & \underline{75.50}±1.96 &\underline{23.17}±1.35  & 36.41±\underline{1.12} & 2.39±1.56   \\
\rowcolor{blue!10} & Ours      & \textbf{77.37}±1.53 & \textbf{26.01}±\textbf{0.75} & \underline{72.72}±\textbf{1.74} & \textbf{76.20}±\underline{1.93} & \textbf{25.77}±1.55 & \textbf{37.69}±1.27 & \underline{1.65}±\underline{0.46}  \\

\midrule
\multirow{7}{*}{\rotatebox[origin=c]{90}{\textbf{AircraftD}}} 
& L2P~\citep{wang2022learning}       &  43.21±1.08 & 20.11±1.20  & 35.52±1.93 & 40.17±2.34 & 21.00±1.21 & 27.21±\underline{1.33} & 15.04±2.10  \\
& DP~\citep{wang2022dualprompt}        &40.18±1.10 & 21.88±1.23 &  34.80±1.79 &42.53±\underline{1.46} & 20.22±\textbf{0.96} & 28.15±1.74 & 12.68±1.74  \\
& GMP~\citep{cendra2025promptccd}       &41.26±\underline{1.04} & 21.12±0.67 &  35.10±2.50 & 32.56±1.75 & 21.25±\underline{1.02} & 26.34±1.48 & 22.65±2.25  \\
& PHE~\citep{zheng2024PHE}  & 45.82±\textbf{1.02} & 22.56±1.05  & \underline{41.72}±2.13 & 47.28±2.88 & 25.09±1.76 & 24.38±1.67 & 7.93±1.13  \\
\cmidrule{2-9}
& HM~\citep{lyu2025test}      & \underline{50.35}±1.08 & \underline{28.01}±0.85 & 40.23±2.22  & \underline{49.77}±1.94 & \underline{29.80}±1.25 & 28.74±1.53 & \underline{5.44}±1.56  \\
& DAA~\citep{liudaa}      & 49.37±1.46 & 27.79±\textbf{0.31} & 40.51±\textbf{1.05} & 48.87±1.73 & 28.37±1.69 & \underline{28.99}±1.69 & 5.61±\underline{1.04}   \\
\rowcolor{blue!10} & Ours     &  \textbf{51.27}±1.51 & \textbf{28.53}±\underline{0.32} & \textbf{41.84}±\underline{1.38} & \textbf{50.84}±\textbf{1.16} & \textbf{30.26}±1.17 & \textbf{30.44}±\textbf{1.23} & \textbf{4.42}±\textbf{0.74} \\
\bottomrule
\end{tabular}
}
\label{tab:major}
\vspace{-10px}
\end{table*}

\subsection{Experimental Details}
\noindent\textbf{Datasets}.
We conduct our experiments based on four benchmark datasets.
CIFAR100 (C100)~\citep{cifar100krizhevsky2009learning} is a commonly used classic dataset. 
Tiny ImageNet~\citep{tinyle2015tiny} is a subset of ImageNet~\citep{deng2009imagenet}.
Caltech-UCSD Birds-200-2011 (CUB)~\citep{cubwah2011caltech} and FGVC-Aircraft~\citep{aircraft2013fine} are complex fine-grained datasets.
All these datasets are split into known and unknown classes (7:3).
The model is trained on the known training set and tested on the mixture of known and unknown test sets.
Because the dataset is used for discovery, we follow the names of four transformed datasets from HM~\citep{lyu2025test}: CIFAR100D, CUB-200D, Tiny-ImageNetD, and AircraftD.

\noindent\textbf{Implementation details}.
In our implementation, we build our method on CLIP~\citep{radford2021clip} as the image and text encoder, which employs a ViT-B/16 backbone~\citep{khan2022transformers} following the pretraining procedure of GCD and TTD work. A linear projection layer is added after the encoders for each modality and only the last layers of image encoders in the backbone are fine-tuned on the known classes, using SGD optimizer and cosine decay learning rate scheduler with an initial learning rate of 0.1 and minimum learning rate of 0.0001, and weight decay of 0.00005. All input images are resized to 224 × 224 and augmented to match the pretrained backbone settings. 

\noindent\textbf{Evaluation metrics}.
Following~\citep{lyu2025test}, we use the following metrics.
For unknown classes, we use two agreement metrics: (1) \textit{True-label Agreement ratio} (\textbf{TA}). This metric measures the maximum proportion of samples from a given true class that are predicted as the same class; and (2) \textit{Cluster Agreement ratio} (\textbf{CA}). This metric measures the maximum proportion of samples from a given predicted cluster that have the same true label.
For known classes, we also employ two key metrics to comprehensively assess the model’s performance: \textit{Known Accuracy} (\textbf{KA}) and \textit{Known Forgetting} (\textbf{KF}). KA measures the traditional classification accuracy of the model on known classes while KF quantifies the degree of performance degradation on known classes over time.
To quantitatively assess the quality of LLM-generated unknown class names, we introduce the \textit{Semantic Score} (\textbf{S$^2$}). This metric evaluates how semantically close the predicted class names are to the ground-truth labels in vector space via a pretrained Sentence-BERT encoder~\citep{reimers2019sentence}, \(\textbf{S}^2(\hat{c}, y) = \frac{f_{\text{BERT}}(\hat{c}) \cdot f_{\text{BERT}}(y)}{\|f_{\text{BERT}}(\hat{c})\|_2 \|f_{\text{BERT}}(y)\|_2}\).

\subsection{Major Comparisons}
In this paper, we first compare our methods with naive thresholding-based training-free methods. When exceeding the threshold, the naive method will be considered to have discovered a new class. And we compare with some training-required methods including L2P~\citep{wang2022learning}, DP~\citep{wang2022dualprompt}, and GMP~\citep{cendra2025promptccd} as these methods update prompts like TTA and CL methods. We also compare with the recent PHE~\citep{zheng2024PHE} method for OCD and HM~\citep{lyu2025test}, DAA~\citep{liudaa} for TTD.

\noindent\textbf{Comparisons on TA \& CA}.
Table~\ref{tab:major} presents a comprehensive comparison of our method against various TTD approaches across four benchmarks. Training‑based methods often suffer from severe performance degradation, indicating that directly updating the model on streaming unknown classes destabilizes known‑class recognition. In contrast, training‑free method like HM achieve better known‑class stability but exhibit only moderate TA, reflecting limited discovery capability. While DAA incorporates a lightweight module that can be updated which still only based on non-semantic discovery and performance remains limited. Notably, a good TTD method must balance and improve both TA and CA; a high value in only one metric often masks poor overall discovery quality. Our proposed SEGA framework achieves the best overall performance, demonstrating that bridging visual clusters to interpretable semantic concepts via LLM‑enriched text alignment improves both discovery accuracy and knowledge retention.

\begin{figure}[htbp]
    \centering    
    \includegraphics[width=1\linewidth]{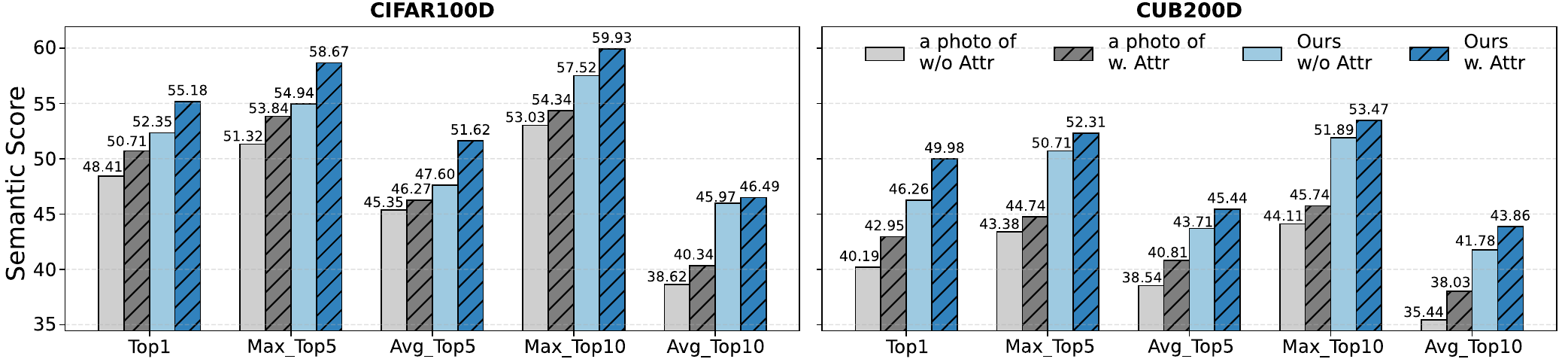}
    \vspace{-18px}
    \caption{
    Comparisons of average Semantic Score (\textbf{S}$^2$) on predictions of unknown classes.
    }
    \label{fig:semantic}
\end{figure}

\noindent\textbf{Comparison on Semantic Score of predictions}. To quantitatively evaluate the semantic quality of our discovered labels, we compute the \textbf{S}$^2$ and Figure~\ref{fig:semantic} reports averaged results over multiple runs on CIFAR100D and CUB200D. First, replacing the simple \textit{“a photo of [CLASS]”} template with our attribute‑rich LLM descriptions method consistently improves Semantic Scores across all metrics, confirming that fine‑grained attribute alignment enhances semantic discriminability. Moreover, when LLM is provided with some attribute assistance, the score of the predicted results is higher, whether it is the template sentence method or our method. Notably, the maximum score among the top‑5 candidates rises over Top‑1, showing that the correct label often appears within the first few LLM suggestions but may not always be ranked first. However, as we enlarge the candidate pool to top‑10, the maximum score only marginally increases, while the average score over top‑10 drops significantly which reflects that LLMs sometimes generate implausible or hallucinated candidates when asked for too many options, and these low‑quality entries pull down the average.

\subsection{Analysis on SEGA}
\begin{figure}[htbp]
    \centering        
    \includegraphics[width=1\linewidth]{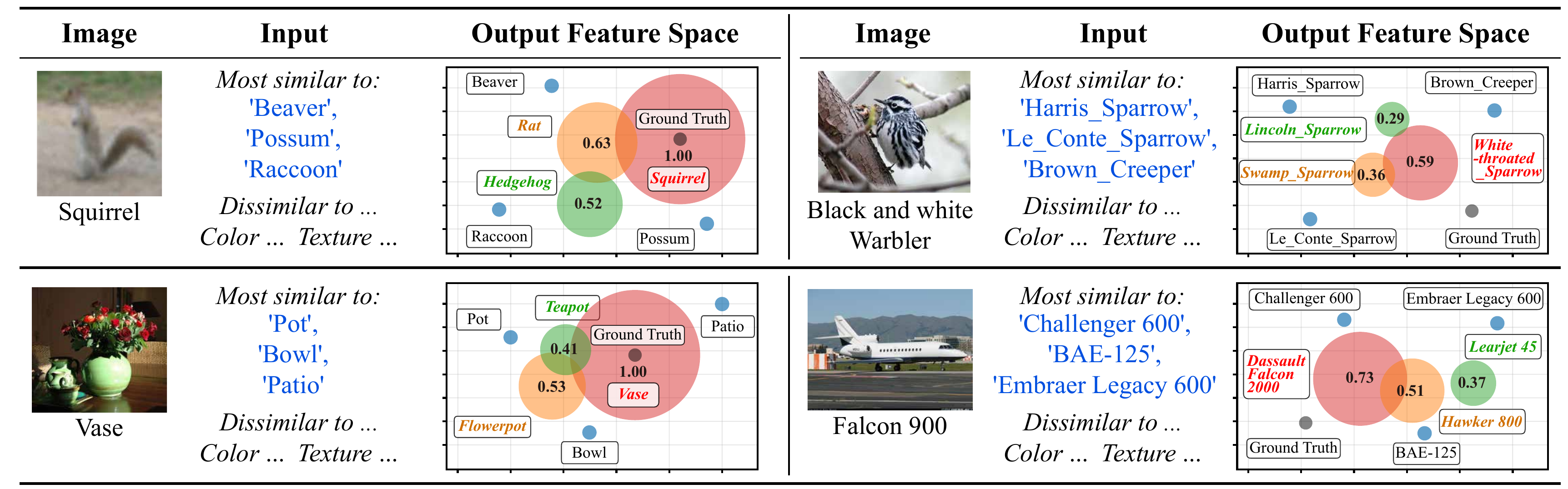}
    \vspace{-18px}
    \caption{
    Some examples of our prediction procedure and textual feature distribution of output. (Size of bubble and value in bubble is \textbf{S}$^2$ of the output.)
    }
    \label{fig:examplemain}
    \vspace{-5px}
\end{figure}

\noindent\textbf{Examples of prediction procedure and performance}. Fig.~\ref{fig:examplemain} showcases representative discovery cases from our framework. When the candidate set includes the ground‑truth label in some cases, the correct label is successfully selected with a high reranking score. Even when exact prediction is not achieved, the top candidate remains semantically close. And from the visualization of the text embedding space, the predicted label consistently lies closer to the ground‑truth embedding than to any known‑class prototype, while still maintaining a clear separation from visually similar known categories. Consequently, even a non‑exact prediction provides a useful anchor for test‑time training which respects both semantic relatedness and discriminability guiding the adaptation for unknown classes without causing catastrophic confusion.

\noindent \textbf{Comparison of compatibility metrics.}
TA and CA directly reflect online discovery quality and class consistency as the primary metrics, while we also provide some clustering metrics that traditional NCD methods use, including Hungarian Cluster Accuracy (HCA)~\citep{meilua2003comparing}, Adjusted Rand Index (ARI)~\citep{rand1971objective}, Normalized Mutual Information (NMI)~\citep{mcdaid2011normalized} and V-Measure~\citep{rosenberg2007v}. The results in Table~\ref{tab:clumetric} demonstrate that our method also has better clustering performance on post evaluation.
\begin{table}[t]
\centering
\setlength{\tabcolsep}{3pt}
\caption{Comparisons of clustering metrics of post evaluation on four datasets.}
\vspace{-10px}
\resizebox{\linewidth}{!}{
\begin{tabular}{l|cccc|cccc|cccc|cccc}
\toprule
\multirow{2}{*}{Method} & \multicolumn{4}{c|}{\textbf{CIFAR100D}}                                & \multicolumn{4}{c|}{\textbf{CUB200D}}                                & \multicolumn{4}{c|}{\textbf{T-ImageNetD}}                                & \multicolumn{4}{c}{\textbf{AircraftD}}                                \\
                        & HCA & ARI & NMI & VM & HCA & ARI & NMI & VM & HCA & ARI & NMI & VM & HCA & ARI & NMI & VM \\
\midrule
HM~\citep{lyu2025test}      & 63.69 & 48.35 & 73.08 & 73.08 & 50.46 & 27.81 & 70.25 & 70.25 &  56.64 & 40.35 & 72.70 & 72.70  & 33.30 & 21.33 & 46.46 & 46.46 \\
DAA~\citep{liudaa}      & 65.62 & 53.96 & 75.20 & 75.21  & 58.19 & 32.43 & 75.29 & 75.29  & 58.91 & 42.50 & 73.53 & 73.53 & 32.08 & 21.20 & 46.44 & 46.44 \\
\rowcolor{blue!10} Ours      & 66.73      & 53.47      & 75.51      & 75.51 & 58.58      & 33.66      & 75.07      & 75.07      & 58.55      & 43.50      & 74.59     & 74.59 & 33.74 & 22.58 & 47.12 & 47.12  \\
\bottomrule
\end{tabular}
}
\vspace{-5px}
\label{tab:clumetric}
\end{table}

\noindent\textbf{Ablation study}. As shown in Table~\ref{tab:ablation}, when none of these components is used which relies solely on a frozen CLIP model, the performance is poor, demonstrating that the pre-trained CLIP model struggles to handle the open-world TTD task due to its limited fine-grained semantic discriminability. Adding only Test-time Training (T) leads to modest improvements, indicating that zero-shot test-time adaptation alone is insufficient without a well-structured semantic space. In contrast, employing only Semantic Pre-construction (C) brings a substantial performance gain, highlighting the importance of constructing an attribute-anchored semantic space before deployment. Finally, the full model combining all three components achieves the best performance, confirming that semantic pre-construction provides a solid foundation, test-time training enables dynamic adaptation, and semantic reranking effectively refines the performance.

\noindent\textbf{Reranking threshold}. Table~\ref{tab:theta} shows the impact of the confidence threshold $\theta$, which determines whether the top-ranked LLM-generated candidate is considered reliable enough to be used as a pseudo label for test-time training. Setting $\theta = 0$ which treats all predictions as reliable leads to the inclusion of many erroneous and noisy pseudo labels, which misguides subsequent test-time training and causes performance degradation. Conversely, setting $\theta = 1$ makes the method degenerate to using only relational cues similar/dissimilar known classes without explicit label and attribute assignment; while this conservative strategy avoids noise, it provides limited guidance and yields suboptimal results. The best performance is achieved at $\theta = 0.8$, where reliable predictions are adopted as pseudo labels to fully leverage semantic guidance, while unreliable ones are discarded and fall back to relation-only prompts, striking an effective balance between noise tolerance and semantic informativeness.
\begin{table}[h]
\vspace{-6px}
	\centering
	\begin{minipage}{0.51\textwidth}
		\centering
		\small
		\makeatletter\def\@captype{table}\makeatother
        \setlength{\tabcolsep}{3pt}
		\caption{Ablation study. C: Semantic Pre-construction, T: Test-time Training, S: Semantic Reranking, Sum: TA+CA. } 
		\label{tab:ablation}
        \vspace{-5px}
        \resizebox{\textwidth}{!}{
		\begin{tabular}{ccc|ccc|cccc}
        \toprule
        \multirow{2}{*}{C}  & \multirow{2}{*}{T}  & \multirow{2}{*}{S} & \multicolumn{3}{c|}{\textbf{Real-time Eval}}                                                                         & \multicolumn{4}{c}{\textbf{Post Eval}}                                                                                                           \\ \cmidrule{4-10} 
            &       &    &  TA            & CA    & Sum        & TA      & CA    & Sum   & KF                        \\
        \midrule
                                                 
                            &                &     & 20.57 & 31.16 & 51.73 & 23.54 & 29.26 & 52.80 & 8.81 \\
  &   $\checkmark$  &   & 21.46 & 40.69 & 62.15 & 24.00 & 33.20 & 57.20 & 11.89 \\

        $\checkmark$       &  &  &  25.07 & 34.59 & 59.65 & 25.27 & 31.53 & 56.80 &  6.54 \\
        $\checkmark$       & $\checkmark$ &  &  31.58 & 42.20 & 73.78 & 32.10 & 35.99 & 68.09 & 4.34 \\
        $\checkmark$       & $\checkmark$ & $\checkmark$ &  34.74 & 47.77 & 82.51 & 38.66 & 39.21 & 77.87 & 3.28 \\
        \bottomrule
        \end{tabular}
         }
	\end{minipage}
    \;
	\begin{minipage}{0.47\textwidth}
		\centering
		\small
		\makeatletter\def\@captype{table}\makeatother
        \setlength{\tabcolsep}{3pt}
    \centering
    \caption{Comparisons of different reranking threshold. Sum: TA+CA.}
    \vspace{-7px}
    \resizebox{1\linewidth}{!}{
    \begin{tabular}{c|ccc|cccc}
        \toprule
         & \multicolumn{3}{c|}{\textbf{Real-time Eval}} & \multicolumn{4}{c}{\textbf{Post Eval}}       \\ 
        \cmidrule{2-8} 
         $\theta$  & TA            & CA    & Sum     & TA       & CA   & Sum  & KF    \\
        \midrule
            0              &  30.88 & 46.59 & 77.46 & 31.40 & 37.57 & 68.96 & 5.39\\
         0.6             & 31.43 & 47.46 & 78.89 & 33.98 & 37.35 & 71.33 & 4.32  \\
         0.7             &  32.35 & 45.50 & 77.85 & 35.27 & 35.93 & 71.19 & 4.06 \\
         0.8             &  34.74 & 47.77 & 82.51 & 38.66 & 39.21 & 77.87 & 3.28 \\
         0.9             &  33.93 & 46.60 & 79.63 & 38.25 & 38.15 & 76.40 & 3.61  \\
        1          & 33.15 & 45.90 & 79.05 & 37.63 & 37.76 & 75.39 & 3.82 \\
        \bottomrule
    \end{tabular}
    }
    \label{tab:theta}
	\end{minipage}
\vspace{-5px}
\end{table}

\begin{figure*}[h]
    \centering    
    \includegraphics[width=\linewidth]{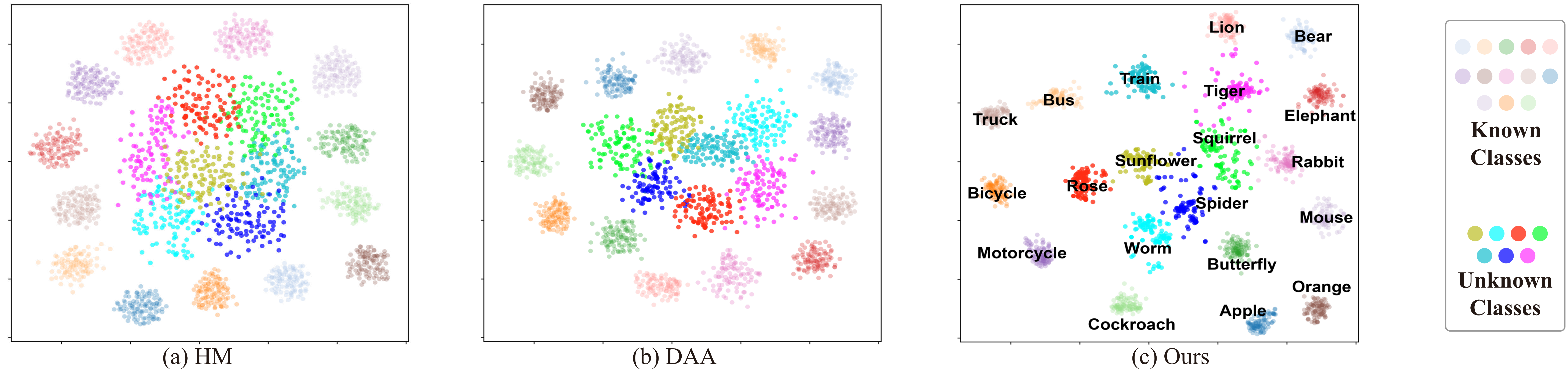}
    \vspace{-18px}
    \caption{
    T-SNE visualization in feature space on CIFAR100D with known and unknown classes.
    }
    \label{fig:tsne}
\end{figure*}

\noindent\textbf{Post visualization using T-SNE}.
In Fig.~\ref{fig:tsne}, we employ t-SNE~\citep{van2008visualizing} to visualize the distribution of test samples. HM produces scattered and overlapping clusters, where known and unknown classes are poorly separated. DAA improves separation by amplifying feature discrepancies, yet the resulting clusters remain largely anonymous and categories are either mixed together or arbitrarily scattered, offering no interpretable organization. Our SEGA yields notably clearer and more structured clusters. Semantically similar classes like Bus and Truck, Apple and Orange, Rose and Sunflower are embedded close to each other, while still maintaining clear inter‑class boundaries not only improving discrimination but also supporting understandable semantics.

\section{Conclusion}
We presented SEGA, a framework that transforms Test-Time Discovery from anonymous clustering into interpretable semantic concept generation. By leveraging LLM-generated attribute descriptions to construct an attribute-anchored visual-semantic space, and introducing Relational Semantic Grounding with confidence-guided test-time stabilization, SEGA enables unknown classes to be named and understood rather than merely separated. Experiments on four benchmarks show consistent improvements over prior TTD methods in both discovery accuracy and semantic coherence. While LLM-generated candidates may occasionally be noisy or hallucinated, the semantic anchoring in a discriminative feature space provides a practical and robust foundation for open-world discovery. Future work will focus on more principled uncertainty estimation and adaptive filtering to further suppress spurious predictions, and to more dynamic, open-ended environments.


\bibliography{main}
\bibliographystyle{bibstyle}

\end{document}